# SWORD: Scene-Grounded Wasserstein Object-Relation Distance for 3D Scene Graph Prediction

U-Chae Jun
wjsdbco@sookmyung.ac.kr
Sookmyung Women's University
Seoul, South Korea

Jaeeun Ko
rhwodms1223@sookmyung.ac.kr
Sookmyung Women's University
Seoul, South Korea

Yoonhyuk Choi*
chldbsgur123@sookmyung.ac.kr
Sookmyung Women's University
Seoul, South Korea

Jiwoo Kang*
jwkang@sookmyung.ac.kr
Sookmyung Women's University
Seoul, South Korea

## Abstract

3D scene graph prediction is commonly optimized with object-node and relation-edge classification losses. These objectives encourage exact object and relation label recovery, but they do not directly optimize semantic proximity to the ground-truth scene graph: two wrong graphs can receive comparable penalties even when one is much closer in object roles and relation contexts. A direct graph-distance loss is difficult in general, since graph matching and graph edit distance require correspondence search and edit-cost design. Our key observation is that prior supervised 3D scene graph prediction protocols are conducted on fixed object-node indices and supervised edge slots. This can turn a scene graph comparison into a well-defined aligned graph-distance problem. To this end, we propose SWORD, a Scene-Grounded Wasserstein Object-Relation Distance that builds block-normalized object and relation ground distance metrics from training scene graph annotations through a deterministic base-to-context order, without external semantic priors. SWORD lifts these metrics to Wasserstein distances over aligned node and null-augmented edge marginals, providing semantic graph-distance supervision beyond exact-label penalties while preserving the ground-truth graph as the unique zero-distance solution in aligned marginal probability space. We prove that SWORD is a strict metric on aligned marginal probabilistic scene graphs, preserves scene-graph-grounded substitution ordering, controls additive graph-grounded semantic observables, and yields a surrogate upper-bounding decoded semantic error.

*Corresponding authors.

## 1 Introduction

3D scene graph prediction represents a 3D scene as a graph of object nodes and relation edges, enabling object-level reasoning beyond isolated recognition. Since 3D semantic scene graphs were introduced for indoor reconstructions, a broad line of work has improved 3D scene graph prediction through RGB-D fusion, point-cloud reasoning, knowledge-guided modeling, visual-linguistic training, multimodal knowledge accumulation, transformers, heterogeneous graph learning, object-centric representations, and edge-centric relation reasoning [1, 6, 10, 11, 22–24, 33–35, 38]. Despite these advances, the supervision objective is still largely based on local label prediction: object nodes are trained to classify object labels, and relation edges are trained to classify relation labels.

Such exact-label supervision preserves the ground-truth label as the target, but it does not define a semantic error geometry among different wrong scene graphs. If the ground-truth object is chair, predicting sofa and predicting wall are both incorrect after hard decoding, even though sofa may be closer to chair in the scene graph context. Likewise, if the ground-truth relation is *near*, predicting adjacent and predicting an opposite directional relation are both wrong, although their semantic severities can differ. Thus, CE/BCE-style node-edge classification optimizes exact-label likelihood, but it does not define an explicit scene-graph-grounded distance that orders semantically near and far wrong predictions.

Fig. 1 illustrates this limitation. The reference graph is compared with three predicted graphs. Under conventional exact-label scoring, graphs (A) and (B) receive almost the same penalties, 0.31 and 0.30, while graph (C) receives a penalty of 0.62, approximately twice as large. This ordering suggests that (A) and (B) are similarly close to the reference and that (C) is substantially farther away. However, a scene-graph-level reading suggests a different ordering. Graph (A) changes object roles and spatial relations in a way that more strongly disrupts the semantic organization of the reference graph. In contrast, graphs (B) and (C) mainly replace objects or relations with alternatives that remain close in the graph context. The example shows that counting local label mismatches can misrepresent the semantic severity of a 3D scene graph prediction error. A desirable loss should therefore preserve the ground-truth graph as the exact optimum, while assigning larger penalties to semantically disruptive substitutions and smaller penalties to graph-grounded alternatives that remain close to the reference.

A natural alternative is graph similarity, but general graph comparison is ill-posed unless one specifies node correspondence, edge correspondence, edit operations, and edit costs. Classical graph edit distance handles this by searching over node and edge insertions, deletions, and substitutions, but the resulting optimization is combinatorial and difficult to use directly as a training objective [2, 5, 7, 26–28]. Our key observation is that many 3D scene graph prediction protocols are not arbitrary graph comparison problems: the predicted graph and the ground-truth graph share object-node indices and supervised relation edge slots [10, 23, 24, 33, 35]. This fixed correspondence turns graph similarity from a matching-dependent problem into a well-posed aligned node-and-edge distance problem.

To this end, we propose **SWORD**, a **S**cene-grounded **W**asserstein **O**bject-**R**elation **D**istance for 3D scene graph prediction. SWORD constructs object and relation ground metrics from training scene graph annotations, using object labels, relation labels, directed edge slots, endpoint contexts, local graph neighborhoods, relation co-occurrences, and typed-degree profiles. Instead of relying on external semantic priors such as language embeddings, knowledge bases, or vision-language models, SWORD derives its semantic geometry from the training scene graphs themselves. To avoid circularity between object and relation semantics, SWORD constructs these costs through a deterministic base-to-context order, from base object and relation metrics to context-aware object and relation metrics. It then uses Wasserstein distances to lift these costs into probability spaces over aligned node and edge marginals [8, 31].

The resulting objective changes the supervision geometry while preserving exact ground-truth optimality. A wrong prediction that is close under the scene-graph-grounded metric can receive a smaller penalty than a semantically distant wrong prediction, but the exact ground-truth graph remains the unique zero-distance solution in probability space. For practical single-label training, we introduce SWORD-Margin, a metric-induced structured log loss that uses SWORD distances as semantic margins. The construction is deterministic once the graph-profile blocks and edge-slot protocol are specified, and it does not require external semantic priors or CE/BCE mixing weights.

The main contributions are summarized as follows. First, we identify fixed correspondence as the structural property that makes semantic graph distance well posed for supervised 3D scene graph prediction, turning graph comparison from a matching-dependent problem into an aligned object-relation distance problem. Second, we construct scene-graph-grounded object and relation metrics from block-normalized training graph profiles, without relying on external semantic priors. Third, we define SWORD as a Wasserstein object-relation graph distance over aligned node and edge marginals, with a null-augmented relation metric for no-relation edge slots. Fourth, we introduce SWORD-Margin for practical single-label training and establish that SWORD is a strict metric on aligned marginal probabilistic scene graphs, preserves scene-graph-grounded substitution ordering, controls additive graph-grounded semantic observables, and has the exact ground-truth graph as the unique zero-distance minimizer.

## 2 Related Work

### 2.1 3D Scene Graph Prediction

3D scene graph prediction is a structured scene understanding task within the field of 3D vision [12–14]. It represents a 3D scene as object nodes and relation edges, and has evolved from indoor reconstruction-based scene graphs to RGB-D, point-cloud, transformer, heterogeneous-graph, object-centric, and edge-centric formulations [6, 10, 22–24, 33–35, 38]. Recent extensions further incorporate visual-linguistic supervision, multimodal knowledge, and pretraining strategies [6, 34]. Across this line of work, the standard prediction interface supervises object nodes and relation edges with annotated labels, and training is typically formulated as independent node and edge classification with CE, BCE, or reweighted classification objectives over object and relation vocabularies [10, 22–24, 33, 35, 38].

These objectives make the ground-truth label optimal, but their label geometry remains exact-match based: they do not specify whether one incorrect object or relation label is semantically closer to the ground truth than another. Consequently, after hard decoding, two wrong labels have the same exact-error status even if one better preserves the scene-graph semantics of the ground-truth scene [6, 22, 24, 33, 34]. SWORD is orthogonal to architectural advances: it exploits the fixed-correspondence node-edge interface to change the supervision geometry from exact-label matching to a scene-graph-grounded object-relation distance built from block-normalized training graph profiles.

### 2.2 Graph Distance and Matching

General graph similarity is difficult because node correspondence, edge correspondence, attribute substitution, and topology changes must be handled jointly [5, 7, 27, 28]. Graph edit distance provides a classical dissimilarity measure based on node and edge insertion, deletion, and substitution, but exact or approximate computation typically involves combinatorial matching procedures such as edit-path search or bipartite matching [2, 7, 26].

The key observation behind SWORD is that this ambiguity disappears in the fixed-correspondence interface used by many 3D scene graph prediction protocols, where predicted and ground-truth graphs share object-node indices and supervised edge slots [10, 23, 24, 33, 35]. This turns graph comparison into a well-posed problem over aligned node and edge marginals, rather than an edit-path or matching problem over arbitrary graph correspondences. Thus, SWORD defines a metric on aligned marginal graph tuples induced by the supervised 3D scene graph prediction interface, not on unaligned graphs.

### 2.3 Wasserstein Ground Metrics

The 1-Wasserstein distance compares probability measures through an explicit ground cost, making it suitable when different label substitutions should have different semantic severities [8, 31]. This differs from standard classification losses, which mainly score the ground-truth label probability without specifying object-relation substitution costs. SWORD uses Wasserstein distances to lift object and relation ground metrics into node-level, edge-level, and graph-level distances over aligned scene graph marginals. The resulting Wasserstein graph distance provides the metric definition, while the cost-augmented log-sum-exp objective used for single-label training is a metric-induced surrogate rather than a separate graph metric.

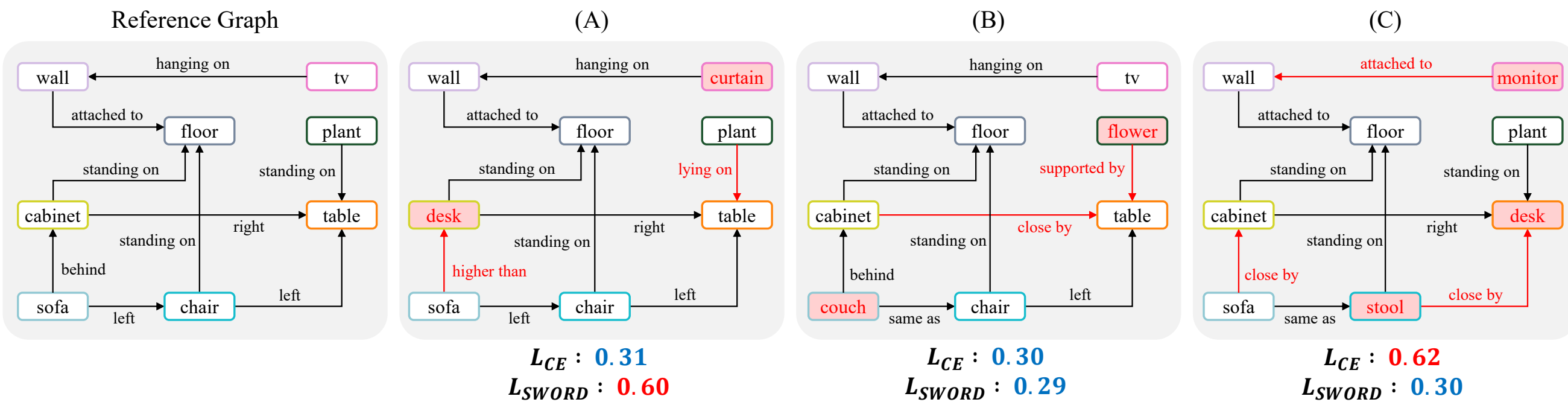

**Figure 1: Existing CE/BCE-style losses mainly reflect local object and relation label mismatches, making (A) and (B) appear similarly close to the reference and (C) about twice as far. A scene-graph-level interpretation suggests that (A) is more disruptive, whereas (B) and (C) remain semantically closer to the reference. $L_{\text{SWORD}}$ previews the graph-grounded ordering by our method.**

Semantic information can also be injected through distributional vector-space semantics, language priors, knowledge bases, vision-language models, or open-vocabulary contrastive learning [3, 4, 6, 16, 21, 30, 34]. SWORD instead constructs its ground metrics from training scene graph annotations themselves, using object-relation contexts, endpoint statistics, local neighborhoods, relation co-occurrences, and typed-degree profiles [23, 24, 33, 35, 38]. This makes the induced semantics dataset-internal and scene-graph-grounded rather than externally imposed. To avoid circularity between object semantics and relation semantics, SWORD constructs these costs through a deterministic base-to-context order, first obtaining base object and relation metrics and then refining them with context-aware object and relation profiles.

## 3 Method

We consider fixed-correspondence 3D scene graph prediction, where object nodes and directed relation edges are supervised by annotated training scene graphs [6, 10, 22–24, 33–35]. Existing CE and BCE objectives penalize node and edge label errors independently, but do not define a graph-level semantic distance to the ground-truth scene graph. SWORD addresses this gap by constructing scene-graph-grounded object and relation metrics from training graph statistics, and lifting them to a Wasserstein distance between aligned probabilistic scene graphs.

### 3.1 3D Scene Graph Estimation Problem

Let the object and relation label spaces be $\mathcal{C}$ and $\mathcal{R}$, and let $\emptyset$ denote the null relation:

$$\mathcal{C} = \{1, \dots, C\}, \qquad \mathcal{R} = \{1, \dots, K\}, \qquad \mathcal{R}_\emptyset = \mathcal{R} \cup \{\emptyset\}. \tag{1}$$

For a supervised scene, the predicted graph and the ground-truth graph share the same object-node indices and directed edge-slot indices:

$$V = \{1, \dots, n\}, \qquad \mathcal{A} \subseteq \{(i, j) : i, j \in V,\ i \neq j\}. \tag{2}$$

The set $\mathcal{A}$ is determined by the benchmark protocol or the prediction architecture, and is not required to contain all ordered pairs. This fixed-correspondence structure is a key observation: general graph comparison requires node correspondences and edit costs, whereas supervised 3D scene graph prediction already provides aligned object-node and edge-slot coordinates. SWORD therefore formulates scene graph comparison as a well-posed metric problem on aligned marginal graph tuples, rather than as a combinatorial graph edit distance or graph matching problem [2, 5, 7, 26, 27].

For node $i$ and edge slot $(i, j)$, the model outputs

$$p_i \in \Delta(\mathcal{C}), \qquad y_i^* \in \mathcal{C}, \tag{3}$$

$$q_{ij} \in \Delta(\mathcal{R}_\emptyset), \qquad r_{ij}^* \in \mathcal{R}_\emptyset \quad \text{for single-label relations,} \tag{4}$$

$$s_{ij}(r) \in [0, 1], \qquad S_{ij}^* \subseteq \mathcal{R} \quad \text{for multi-label relations.} \tag{5}$$

An aligned marginal probabilistic scene graph is represented as

$$G = \left(\{p_i^G\}_{i \in V}, \{\mu_{ij}^G\}_{(i,j) \in \mathcal{A}}\right), \qquad G^* = \left(\{\delta_{y_i^*}\}_{i \in V}, \{\mu_{ij}^{\text{gt}}\}_{(i,j) \in \mathcal{A}}\right). \tag{6}$$

SWORD is a metric on aligned marginal graph tuples of the form in (6). It is not a metric over arbitrary unaligned graphs or unspecified full joint graph distributions.

For finite distributions over a token space, SWORD uses total variation when no lower-level ground metric is defined, and Wasserstein distance otherwise:

$$d_{\text{TV}}(P, Q) = \frac{1}{2} \sum_{x \in \mathcal{X}} |P(x) - Q(x)|, \tag{7}$$

$$W_1(\mu, \nu; d) = \min_{\pi \in \Pi(\mu, \nu)} \sum_{x, y \in \mathcal{X}} \pi(x, y) d(x, y). \tag{8}$$

When $d$ is a metric, $W_1(\cdot, \cdot; d)$ is a metric on probability measures over $\mathcal{X}$ [8, 31].

### 3.2 Scene-Graph-Grounded Ground Metrics

SWORD constructs object and relation ground metrics only from training scene graph annotations: object labels, relation labels, directed incidence, endpoint contexts, local neighborhoods, typed degrees, and relation co-occurrence. It does not use raw 3D coordinates, geometry, shape descriptors, point-cloud features, language priors, or vision-language priors. Thus, the induced semantics are dataset-internal and scene-graph-grounded.

Object and relation semantics are mutually dependent in a scene graph. Object labels are characterized by their relation and neighbor contexts, while relation labels are characterized by endpoint objects and local relation contexts. To avoid a circular definition, SWORD uses an acyclic base-to-context construction:

$$d_O^{\text{base}} \longrightarrow d_R^{\text{base}} \longrightarrow d_O \longrightarrow d_R. \tag{9}$$

**Empirical graph profiles.** For a graph-profile channel $m$ with finite token space $\mathcal{X}_m$, let $\mathcal{I}_{\ell,m}$ be the set of training occurrences associated with label $\ell$, and let $\phi_m(\alpha)$ be the token extracted from occurrence $\alpha$. A channel-specific empty token $\varnothing_m$ is used when no occurrence is observed. The empirical profile is

$$\widehat{P}_{\ell,m}(x) = \begin{cases} \frac{1}{|\mathcal{I}_{\ell,m}|}\sum_{\alpha\in\mathcal{I}_{\ell,m}} \mathbf{1}[\phi_m(\alpha)=x], & |\mathcal{I}_{\ell,m}|>0, \\ \mathbf{1}[x=\varnothing_m], & |\mathcal{I}_{\ell,m}|=0. \end{cases} \tag{10}$$

Channels are grouped into semantically homogeneous blocks so that no profile family dominates merely by containing more channels. For block family $\mathcal{B}_t$ at construction stage $t$, define

$$\text{BA}_{\mathcal{B}_t}\left[\{A_m\}_m\right](\ell,\ell') = \frac{1}{|\mathcal{B}_t|}\sum_{B\in\mathcal{B}_t}\frac{1}{|\mathcal{M}_B|}\sum_{m\in\mathcal{M}_B} A_m(\ell,\ell'). \tag{11}$$

The object blocks include neighbor-object, incoming-relation, outgoing-relation, relation-neighbor, and typed-degree profiles. The relation blocks include subject-class, object-class, subject-object pair, reverse-relation, and local relation co-occurrence profiles. These block families are fixed before training and are part of the method specification.

When a Wasserstein channel contains $\varnothing_m$, a non-empty ground metric $d_m^+$ with diameter at most one is extended as

$$d_m^{\varnothing}(x,y) = \begin{cases} d_m^+(x,y), & x,y\neq\varnothing_m, \\ 1, & x\neq\varnothing_m,\ y=\varnothing_m, \\ 1, & x=\varnothing_m,\ y\neq\varnothing_m, \\ 0, & x=y=\varnothing_m. \end{cases} \tag{12}$$

**Strict correction.** Empirical profile distances may be pseudometrics because distinct labels can have identical graph profiles. For a finite label space $\mathcal{L}$ and raw pseudometric $d_L^0$, define

$$\Delta_L = \max_{u,v\in\mathcal{L}} d_L^0(u,v), \qquad \bar{d}_L^0(u,v) = \begin{cases} d_L^0(u,v)/\Delta_L, & \Delta_L>0, \\ 0, & \Delta_L=0, \end{cases}$$

$$\mathcal{P}_L = \{\bar{d}_L^0(u,v) : u\neq v,\ \bar{d}_L^0(u,v)>0\}, \qquad \eta_L = \begin{cases} \frac{1}{2}\min\mathcal{P}_L, & \mathcal{P}_L\neq\emptyset, \\ 1, & \mathcal{P}_L=\emptyset, \end{cases}$$

$$\text{SC}_{\mathcal{L}}[d_L^0](u,v) = \frac{\max\left(\bar{d}_L^0(u,v),\eta_L\mathbf{1}[u\neq v]\right)}{\max_{a,b\in\mathcal{L}}\max\left(\bar{d}_L^0(a,b),\eta_L\mathbf{1}[a\neq b]\right)}. \tag{13}$$

The scale $\eta_L$ is determined by the training-set distance matrix and is not tuned on validation data. It separates only label pairs indistinguishable under the chosen profiles, without changing the ordering of pairs already separated by nonzero graph-grounded distance.

**Base metrics.** The base object metric compares exact graph-token profiles by total variation:

$$A_m^{O,\text{base}}(a,b) = d_{\text{TV}}\left(\widehat{P}_{a,m},\widehat{P}_{b,m}\right), \tag{14}$$
$$d_{O,\text{base}}^0(a,b) = \text{BA}_{\mathcal{B}_O^{\text{base}}}\left[\{A_m^{O,\text{base}}\}_m\right](a,b), \qquad d_O^{\text{base}} = \text{SC}_C[d_{O,\text{base}}^0].$$

Using $d_O^{\text{base}}$, define a base endpoint-pair metric and the base relation metric:

$$d_{\text{pair}}^{\text{base}}\big((a,b),(a',b')\big) = \frac{d_O^{\text{base}}(a,a')+d_O^{\text{base}}(b,b')}{2}, \tag{15}$$

$$A_m^{R,\text{base}}(r,s) = D_m^{R,\text{base}}\left(\widehat{P}_{r,m},\widehat{P}_{s,m}\right), \tag{16}$$
$$d_{R,\text{base}}^0(r,s) = \text{BA}_{\mathcal{B}_R^{\text{base}}}\left[\{A_m^{R,\text{base}}\}_m\right](r,s),\ \ d_R^{\text{base}} = \text{SC}_{\mathcal{R}}[d_{R,\text{base}}^0].$$

Here subject and object channels use $W_1(\cdot,\cdot;d_O^{\text{base}})$, endpoint-pair channels use $W_1(\cdot,\cdot;d_{\text{pair}}^{\text{base}})$, and exact relation-token channels use $d_{\text{TV}}$.

**Context-aware metrics.** The context-aware object metric compares object profiles using the base object and relation metrics as ground costs. First define

$$d_{\text{rel-nbr}}^{\text{base}}\big((r,c),(r',c')\big) = \frac{d_R^{\text{base}}(r,r')+d_O^{\text{base}}(c,c')}{2}. \tag{17}$$

The context-aware object metric is

$$A_m^{O,\text{ctx}}(a,b) = D_m^O\left(\widehat{P}_{a,m},\widehat{P}_{b,m}\right),$$
$$d_O^0(a,b) = \text{BA}_{\mathcal{B}_O^{\text{ctx}}}\left[\{A_m^{O,\text{ctx}}\}_m\right](a,b), \qquad d_O = \text{SC}_C[d_O^0]. \tag{18}$$

In (18), neighbor-object channels use $W_1(\cdot,\cdot;d_O^{\text{base}})$, relation-label channels use $W_1(\cdot,\cdot;d_R^{\text{base}})$, relation-neighbor channels use $W_1(\cdot,\cdot;d_{\text{rel-nbr}}^{\text{base}})$, and typed-degree channels use $d_{\text{TV}}$.

Finally, the context-aware relation metric uses the object metric $d_O$ for endpoint comparison and $d_R^{\text{base}}$ for relation-token comparison:

$$d_{\text{pair}}\big((a,b),(a',b')\big) = \frac{d_O(a,a')+d_O(b,b')}{2}, \tag{19}$$

$$A_m^{R,\text{ctx}}(r,s) = D_m^R\left(\widehat{P}_{r,m},\widehat{P}_{s,m}\right),$$
$$d_R^0(r,s) = \text{BA}_{\mathcal{B}_R^{\text{ctx}}}\left[\{A_m^{R,\text{ctx}}\}_m\right](r,s), \qquad d_R = \text{SC}_{\mathcal{R}}[d_R^0]. \tag{20}$$

Subject and object channels use $W_1(\cdot,\cdot;d_O)$, endpoint-pair channels use $W_1(\cdot,\cdot;d_{\text{pair}})$, and reverse-relation and co-occurrence channels use $W_1(\cdot,\cdot;d_R^{\text{base}})$. The construction is deterministic and non-iterative.

**Null augmentation.** Since $d_R$ has normalized diameter one, we extend it to $\mathcal{R}_\emptyset$ by

$$d_R^{\emptyset}(x,y) = \begin{cases} d_R(x,y), & x,y\in\mathcal{R}, \\ 1, & x\in\mathcal{R},\ y=\emptyset, \\ 1, & x=\emptyset,\ y\in\mathcal{R}, \\ 0, & x=y=\emptyset. \end{cases} \tag{21}$$

The null cost is tied to the normalized diameter of $d_R$ and is not selected by validation as an insertion or deletion cost.

## 3.3 SWORD Distance and Training Objectives

For a node $i$, the SWORD node atom distance to the ground truth is

$$\ell_i^V = W_1(p_i,\delta_{y_i^*};d_O) = \sum_{c\in C} p_i(c)\,d_O(c,y_i^*). \tag{22}$$

For a single-label edge slot $(i,j)$, the edge atom distance is

$$\ell_{ij}^E = W_1(q_{ij},\delta_{r_{ij}^*};d_R^{\emptyset}) = \sum_{r\in\mathcal{R}_\emptyset} q_{ij}(r)\,d_R^{\emptyset}(r,r_{ij}^*). \tag{23}$$

For multi-label relation prediction, the sigmoid vector and ground-truth relation set are converted into probability measures on $\mathcal{R}_\emptyset$:

$$\mu_{ij}^{\text{pred}} = \frac{1}{K}\sum_{r\in\mathcal{R}} s_{ij}(r)\delta_r + \left(1 - \frac{1}{K}\sum_{r\in\mathcal{R}} s_{ij}(r)\right)\delta_\emptyset, \tag{24}$$

$$\mu_{ij}^{\text{gt}} = \frac{1}{K}\sum_{r\in S_{ij}^*} \delta_r + \left(1 - \frac{|S_{ij}^*|}{K}\right)\delta_\emptyset, \tag{25}$$

$$\ell_{ij}^{E,\text{multi}} = K \cdot W_1\left(\mu_{ij}^{\text{pred}}, \mu_{ij}^{\text{gt}}; d_R^\emptyset\right). \tag{26}$$

The factor $K$ is fixed by the relation vocabulary size and compensates for the $1/K$ mass assigned to each real relation slot. For the graph-level notation below, $\mu_{ij}$ denotes the edge marginal used by SWORD. In the single-label setting, $\mu_{ij}^{\text{pred}} = q_{ij}$ and $\mu_{ij}^{\text{gt}} = \delta_{r_{ij}^*}$. In the multi-label setting, $\mu_{ij}^{\text{pred}}$ and $\mu_{ij}^{\text{gt}}$ are the null-augmented measures defined in Eqs. (24) and (25).

For two aligned marginal scene graphs $G$ and $H$, define

$$\beta = \begin{cases} 1, & \text{single-label relation setting,} \\ K, & \text{multi-label relation setting,} \end{cases} \tag{27}$$

$$D_\beta(G,H) = \frac{\sum_{i\in V} W_1(p_i^G, p_i^H; d_O) + \beta\sum_{(i,j)\in\mathcal{A}} W_1(\mu_{ij}^G, \mu_{ij}^H; d_R^\emptyset)}{|V| + \beta|\mathcal{A}|}. \tag{28}$$

The pure SWORD loss is

$$\mathcal{L}_{\text{SWORD}}(\hat{G}, G^*) = D_\beta(\hat{G}, G^*). \tag{29}$$

For single-label logit training, Wasserstein-to-Dirac losses are bounded linear functions of predicted probabilities and may give weak gradients for confident wrong predictions. We therefore use a cost-augmented log-sum-exp surrogate:

$$\ell_i^{V,\text{mar}} = \log\sum_{c\in C}\exp\left(z_i(c) + d_O(c, y_i^*)\right) - z_i(y_i^*), \tag{30}$$

$$\ell_{ij}^{E,\text{mar}} = \log\sum_{r\in\mathcal{R}_\emptyset}\exp\left(u_{ij}(r) + d_R^\emptyset(r, r_{ij}^*)\right) - u_{ij}(r_{ij}^*), \tag{31}$$

$$\mathcal{L}_{\text{SWORD-Margin}} = \frac{\sum_{i\in V}\ell_i^{V,\text{mar}} + \sum_{(i,j)\in\mathcal{A}}\ell_{ij}^{E,\text{mar}}}{|V| + |\mathcal{A}|}. \tag{32}$$

SWORD-Margin is not itself a metric on probabilistic scene graphs; it is a metric-induced structured log loss whose semantic margins are exactly the SWORD ground distances. This cost-augmented form is related to large-margin and softmax-margin structured prediction, but the margin values here are induced by scene-graph-grounded object and relation metrics rather than by hand-designed task costs [9, 29]. For multi-label relations, we use

$$\mathcal{L}_{\text{SWORD}}^{\text{multi}} = \frac{\sum_{i\in V}\ell_i^{V,\text{mar}} + \sum_{(i,j)\in\mathcal{A}}\ell_{ij}^{E,\text{multi}}}{|V| + K|\mathcal{A}|}. \tag{33}$$

In Eq. (33), the edge term is the multi-label Wasserstein relation loss, while the object term uses the same SWORD-Margin surrogate as in Eq. (30), because object prediction remains single-label in both relation protocols.

## 3.4 Theoretical Properties

**Proposition 1. Ground distances are strict metrics.** The distances $d_O$ and $d_R$ in (18) and (20) are strict metrics on $C$ and $\mathcal{R}$, respectively.

*Proof.* Total variation is a metric on finite categorical distributions, and Wasserstein distance is a metric when its ground cost is a metric. The empty-token extension in (12) is a metric because the non-empty diameter is at most one and the empty token is placed at distance one from all non-empty tokens. Therefore, all profile distances used in the acyclic order of (9) induce pseudometrics on label spaces, and the block-normalized averages in (11) preserve pseudometricity as positive weighted sums. The strict correction in (13) takes the pointwise maximum with a scaled discrete metric, which preserves the triangle inequality:

$$\begin{aligned}\max(d(x,z), e(x,z)) &\le \max(d(x,y) + d(y,z), e(x,y) + e(y,z)) \\ &\le \max(d(x,y), e(x,y)) + \max(d(y,z), e(y,z)). \end{aligned} \tag{34}$$

It also gives positive separation to all distinct labels, and the final normalization preserves metric properties. Hence $d_O$ and $d_R$ are strict metrics. □

**Proposition 2. The null-augmented relation distance is a metric.** The distance $d_R^\emptyset$ in (21) is a metric on $\mathcal{R}_\emptyset$.

*Proof.* For real relations, the claim follows from Proposition 1. For any $r, s \in \mathcal{R}$, the only nontrivial triangle involving $\emptyset$ is

$$d_R^\emptyset(r,s) = d_R(r,s) \le 1 \le d_R^\emptyset(r,\emptyset) + d_R^\emptyset(\emptyset, s). \tag{35}$$

All other cases are immediate from (21). □

**Theorem 1. SWORD is a strict metric on aligned marginal scene graphs.** For fixed $V$ and $\mathcal{A}$, $D_\beta$ in (28) is a strict metric on aligned marginal graph tuples.

*Proof.* By Propositions 1 and 2, $d_O$ and $d_R^\emptyset$ are strict metrics. Therefore, each Wasserstein term in (28) is a metric on the corresponding probability simplex. A positive weighted sum of metrics followed by positive normalization is a metric. If $D_\beta(G,H) = 0$, every aligned node marginal and every aligned edge marginal coincide:

$$p_i^G = p_i^H \quad \forall i \in V, \qquad \mu_{ij}^G = \mu_{ij}^H \quad \forall (i,j) \in \mathcal{A}. \tag{36}$$

Under fixed correspondence, this is equality of aligned marginal scene graph tuples. □

**Theorem 2. SWORD preserves scene-graph-grounded substitution ordering.** Let $G^*$ be the ground-truth graph. Suppose hard graphs $H_a$ and $H_b$ differ from $G^*$ only at node $i$, where they assign object labels $a$ and $b$, respectively. If

$$d_O(a, y_i^*) < d_O(b, y_i^*), \tag{37}$$

then

$$D_\beta(H_a, G^*) < D_\beta(H_b, G^*). \tag{38}$$

*Proof.* All graph coordinates cancel except node $i$:

$$D_\beta(H_a, G^*) - D_\beta(H_b, G^*) = \frac{d_O(a, y_i^*) - d_O(b, y_i^*)}{|V| + \beta|\mathcal{A}|}. \tag{39}$$

The right-hand side is negative by (37). The relation case is identical with $d_R^\emptyset$ replacing $d_O$. □

**Theorem 3. SWORD controls graph-grounded semantic observables.** Let $\varphi_i : C \to \mathbb{R}$ be 1-Lipschitz with respect to $d_O$, and let $\psi_{ij} : \mathcal{R}_\emptyset \to \mathbb{R}$ be 1-Lipschitz with respect to $d_R^\emptyset$. Define

$$\Psi(G) = \frac{\sum_{i\in V}\mathbb{E}_{c\sim p_i^G}[\varphi_i(c)] + \beta\sum_{(i,j)\in\mathcal{A}}\mathbb{E}_{r\sim\mu_{ij}^G}[\psi_{ij}(r)]}{|V| + \beta|\mathcal{A}|}. \tag{40}$$

Then, for any aligned marginal graphs $G$ and $H$,

$$|\Psi(G) - \Psi(H)| \le D_\beta(G,H). \tag{41}$$

*Proof.* By the Kantorovich-Rubinstein dual characterization of $W_1$, the expectation difference of any 1-Lipschitz observable is bounded by the corresponding Wasserstein distance. Applying this coordinate-wise to all nodes and edges, summing, and normalizing gives (41). □

**Corollary 1. The ground-truth graph is the unique zero-distance minimizer.** For the pure SWORD loss,

$$\mathcal{L}_{\text{SWORD}}(\hat{G}, G^*) = 0 \iff \hat{G} = G^* \tag{42}$$

in the aligned marginal probability space. With finite softmax or sigmoid logits, exact zero may only be approached in the infinite-logit limit, as in CE and BCE.

**Proposition 3. SWORD-Margin upper-bounds decoded semantic error.** Let

$$\hat{y}_i = \arg\max_{c \in C} z_i(c), \qquad \hat{r}_{ij} = \arg\max_{r \in \mathcal{R}_\emptyset} u_{ij}(r). \tag{43}$$

Then

$$\ell_i^{V,\text{mar}} \geq d_O(\hat{y}_i, y_i^*), \qquad \ell_{ij}^{E,\text{mar}} \geq d_R^\emptyset(\hat{r}_{ij}, r_{ij}^*). \tag{44}$$

*Proof.* The log-sum-exp term is at least the decoded-class term, and the decoded logit is no smaller than the ground-truth logit. Applying this to (30) and (31) gives (44). □

SWORD does not claim to recover universal human commonsense similarity. Its semantics are dataset-internal and graph-grounded. Theorems 2 and 3 show that the ordering and Lipschitz stability induced by the constructed object and relation metrics are preserved at the aligned graph-distance level for single-coordinate substitutions and additive aligned node-edge observables.

## 3.5 Implementation and Complexity

SWORD separates offline metric construction from online training. Offline, it extracts graph profiles from the training set, computes $d_O^{\text{base}}$, $d_R^{\text{base}}$, $d_O$, $d_R$, and $d_R^\emptyset$, then stores the object and relation cost matrices. Online, the loss uses only cost-matrix lookup and standard probability or log-sum-exp operations.

Let $M_O$ and $M_R$ denote the total numbers of object and relation graph-profile channels across all blocks, and let $m = |\mathcal{A}|$. The one-time precomputation and online costs are

$$T_{\text{pre}} = O(C^2 M_O + K^2 M_R), \qquad M_{\text{store}} = O(C^2 + (K+1)^2), \tag{45}$$

$$T_{\text{single}} = O(nC + mK), \qquad T_{\text{multi}} = O(nC + mK^3). \tag{46}$$

Thus, in the single-label setting, SWORD-Margin has the same asymptotic order as standard softmax node and relation classification, with additional fixed semantic cost lookup.

Once the graph-profile block families and benchmark edge-slot protocol are specified before validation, SWORD introduces no validation-tuned loss-side scalar weight, relation insertion cost, relation deletion cost, margin scale, temperature, Sinkhorn coefficient, learned metric parameter, or external-prior weight. The block-normalized averaging, strict correction, null cost, and multi-label scaling are fixed by (11), (13), (21), and (26), respectively.

# 4 Experiments

## 4.1 Experimental Setting

*4.1.1 Dataset.* We conducted experiments on the 3DSSG dataset [33], a semantically enriched 3D scene graph dataset built upon 3RScan [32]. 3DSSG provides object instances, semantic object categories, and directed object-object relation annotations for real-world indoor RGB-D reconstructions. We used two label settings that are commonly adopted in recent 3D scene graph prediction studies.

For the multi-label setting, following recent works [10, 34], we used the original 3DSSG label space with 160 object categories and 26 predicate types, together with the standard train/validation split over 1,553 indoor scenes. In this setting, an ordered object pair may be associated with multiple relation labels.

For the single-label setting, we used the 3DSSG-O27R16 label space and train/validation split introduced by SGGPoint [37]. This setting is based on the RIO27 object annotation and maps the original object vocabulary into 27 object classes and the relation vocabulary into 16 relation classes. The detailed single-label conversion and scoring protocol are described in Sec. 4.1.2, since they affect how predictions are evaluated rather than the source dataset itself.

*4.1.2 Metrics and Task Description.* Because the multi-label and single-label settings use different label spaces and scoring rules, we report their results separately.

*Multi-label Setting.* We follow the standard 3DSSG evaluation protocol [33, 38]. We evaluate object, predicate, and triplet prediction using top-$K$ Recall (R@$K$) and top-$K$ mean Recall (mR@$K$). For triplet prediction, a predicted triplet is counted as correct only when its subject category, predicate label, and object category all match a ground-truth triplet. We report results on two standard scene graph prediction tasks [36]: Predicate Classification (PredCls) and Scene Graph Classification (SGCls). In PredCls, ground-truth object instances and object categories are given, and the model predicts predicates for directed object pairs. In SGCls, ground-truth object instances are given, and the model predicts both object categories and predicates, so the final subject-predicate-object triplets are evaluated.

*Single-label Setting.* We follow the 3DSSG-O27R16 protocol used in recent single-label methods [22, 37], where comparative predicates are removed, directional proximity predicates are merged into a unified spatial proximity relation, and ambiguous multi-label relation annotations are converted into a single target relation label. We evaluate object prediction and relationship prediction with R@$K$, and predicate prediction with F1@$K$. Unlike recall alone, F1@$K$ jointly reflects precision and recall, and therefore penalizes both false positive predicate predictions and missed predicate labels. This is especially relevant in the single-label setting, where each directed object pair is represented by one relation target.

These automatic metrics are based on exact label agreement. They are therefore appropriate for comparison with prior work, but they do not directly measure whether an incorrect label is semantically close to the ground truth or whether the predicted scene graph remains plausible as a whole. We thus complement the quantitative benchmark results with user studies in Sec. 4.4.

*4.1.3 Implementation Details.* All experiments were conducted on a single NVIDIA RTX 6000 Ada GPU with 48GB memory. We trained all models for 100 epochs using the AdamW optimizer [15, 20] with a global batch size of 8 scenes and an initial learning rate of $1 \times 10^{-4}$. The learning rate was decayed with a cosine schedule [19]. For each relation setting, we followed the original implementation of the corresponding backbone in terms of architecture, input preprocessing, training configuration and inference pipeline, while replacing only the supervision objective with the proposed SWORD objective. The object and relation distance matrices were computed once from the training split and kept fixed during training. During training, the precomputed matrices were used only through distance lookup in the corresponding SWORD training objective, either as Wasserstein ground costs or as semantic cost margins.

**Table 1: Object, predicate, and triplet recall on 3DSSG in the multi-label setting.**

| Model | Object | | | | Predicate | | | | Triplet | | | |
|---|---|---|---|---|---|---|---|---|---|---|---|---|
| | R@1 | R@5 | mR@1 | mR@5 | R@1 | R@3 | mR@1 | mR@3 | R@50 | R@100 | mR@50 | mR@100 |
| SGPN [33] | 49.46 | 73.99 | 10.86 | 27.54 | 86.92 | 94.76 | 32.01 | 55.22 | 85.38 | 88.59 | 41.52 | 51.92 |
| SGFN [35] | 53.36 | 76.88 | 21.24 | 46.68 | 89.00 | 97.71 | 41.89 | 70.82 | 88.59 | 91.14 | 58.37 | 67.61 |
| VL-SAT [34] | 55.93 | 78.06 | 21.41 | 46.14 | 89.81 | 98.46 | 54.03 | 77.67 | 89.35 | 92.20 | 65.09 | 73.59 |
| OCRL [10] | 59.53 | 81.20 | 22.55 | 48.10 | 91.27 | 98.48 | 56.32 | 76.26 | 91.40 | 93.80 | 65.31 | 74.54 |
| **SWORD** | **59.86** | **82.58** | **24.08** | **50.84** | **91.38** | **99.02** | **58.74** | **78.92** | **92.36** | **94.82** | **68.41** | **78.06** |

**Table 2: PredCls and SGCls results on 3DSSG in the multi-label setting.**

| Model | PredCls | | | | | | SGCls | | | | | |
|---|---|---|---|---|---|---|---|---|---|---|---|---|
| | R@20 | R@50 | R@100 | mR@20 | mR@50 | mR@100 | R@20 | R@50 | R@100 | mR@20 | mR@50 | mR@100 |
| SGPN [33] | 51.9 | 58.0 | 58.5 | 32.1 | 38.4 | 38.9 | 27.0 | 28.8 | 29.0 | 19.7 | 22.6 | 23.1 |
| SGFN [35] | 65.9 | 78.8 | 79.6 | 46.1 | 54.8 | 55.1 | 29.5 | 31.2 | 31.2 | 20.5 | 23.1 | 23.1 |
| VL-SAT [34] | 67.8 | 79.9 | 80.8 | 57.8 | 64.2 | 64.3 | 32.0 | 33.5 | 33.7 | 31.0 | 32.6 | 32.7 |
| OCRL [10] | 70.2 | 82.0 | 82.6 | 58.8 | 66.1 | 66.4 | 36.1 | 37.7 | 37.8 | 35.3 | 37.7 | 37.9 |
| **SWORD** | **71.6** | **83.5** | **84.1** | **61.4** | **69.0** | **69.4** | **37.2** | **39.0** | **39.2** | **37.1** | **39.7** | **40.1** |

## 4.2 Comparison with Related Methods

We evaluated SWORD through comparisons with recent 3D scene graph prediction methods. Since SWORD supports both multi-label and single-label relation prediction, we reported the results separately for the two settings.

*4.2.1 Multi-Label Setting.* For the multi-label setting, we compared SWORD with SGPN [33], SGFN [35], VL-SAT [34], and OCRL [10]. To isolate the effect of the proposed supervision, we used OCRL as the backbone and replaced only the loss-side training objective with the multi-label SWORD objective in Eq. (33). The input modality, network architecture, graph construction, and evaluation protocol are kept unchanged. Thus, this comparison directly reflects the benefit of introducing scene-graph-grounded object and relation costs into training.

Table 1 reports object, predicate, and triplet prediction results. Among the compared methods under the same multi-label evaluation protocol, SWORD achieves the strongest performance across all reported metrics. This is notable because the evaluation still uses exact ground-truth labels: SWORD does not relax the test criterion, but improves strict label ranking by changing the geometry of the training signal. Compared with OCRL, the most meaningful trend appears in mean-recall and triplet-level measures. Mean recall is less dominated by frequent classes, so the consistent improvement suggests that SWORD provides more informative supervision for rare, ambiguous, or visually confusable object-relation patterns. This behavior is consistent with the design of SWORD: semantically nearby substitutions receive smaller graph-grounded costs than distant substitutions, while the underlying SWORD distance still preserves the ground-truth graph as the unique zero-distance solution. As a result, the model can organize its prediction distribution around plausible semantic neighborhoods without giving up exact-label optimality. The triplet results further show that the improvement is not limited to isolated object or predicate classification. Since a triplet is correct only when the subject, predicate, and object are jointly aligned, the strong triplet performance indicates that SWORD improves the compatibility of node and edge predictions at the graph level. This supports the central motivation of SWORD: object and relation labels should not be supervised as independent local decisions only, but should be structured by their scene-graph-grounded semantic proximity.

Table 2 shows the PredCls and SGCls results. In PredCls, object categories are given, so the task emphasizes relation reasoning. SWORD consistently outperforms OCRL, indicating that the proposed relation distance improves predicate prediction itself rather than only benefiting from better object recognition. In SGCls, both object classification and relation prediction are involved, and SWORD again maintains the best overall performance. This trend suggests that the object and relation ground metrics work coherently: the object-side margin improves node predictions, while the multi-label Wasserstein relation loss provides structured supervision over predicate labels and the null relation.

Figure 2 provides qualitative evidence for this behavior. In the bathroom scene, OCRL introduces objects such as *microwave* and *toilet*, and produces relations such as *part of* and *behind* that are less consistent with the visible scene structure. SWORD still makes some exact-label mismatches, such as predicting *bath cabinet* or *front*, but these predictions remain closer to the bathroom context and preserve most of the reference object-relation layout. In the desk scene, OCRL replaces scene-consistent objects with *tv* and *monitor* and assigns an implausible *hanging on* relation to the table. In contrast, SWORD preserves the window-lamp-table-chair arrangement and predicts a physically compatible support relation between the lamp and the table. These examples illustrate that SWORD tends to avoid semantically distant substitutions and yields more coherent graphs when exact labels are uncertain.

Overall, the multi-label results demonstrate that SWORD improves both strict quantitative metrics and qualitative graph coherence. By deriving object and relation distances from training scene graphs and using them as supervision geometry, SWORD encourages predictions that are not only more accurate under exact-label evaluation but also more consistent with the semantic and structural organization of 3D scenes.

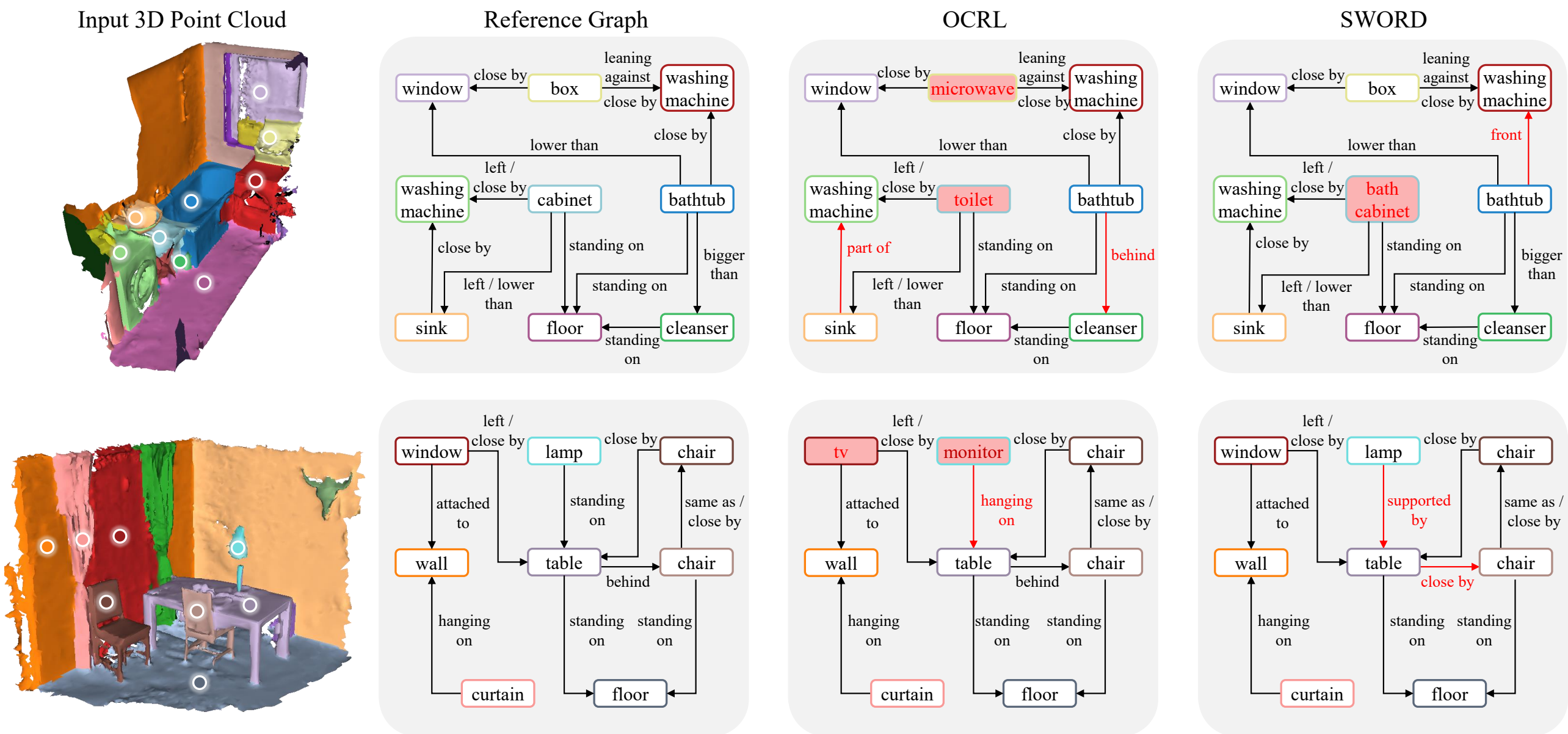

**Figure 2: Qualitative comparison in the multi-label setting. Red nodes and edges denote deviations from the reference graph.**

**Table 3: Quantitative comparison under the standard single-label protocol [22, 37].**

| Model | Object | | Predicate | | Relationship | |
|---|---|---|---|---|---|---|
| | R@5 | R@10 | F1@3 | F1@5 | R@50 | R@100 |
| SGGPoint [37] | 90.70 | 97.58 | 78.88 | 90.86 | 39.91 | 48.68 |
| SGFormer [22] | 96.41 | 98.48 | 85.12 | 91.58 | 56.25 | 60.67 |
| SWORD (Multi) | 96.48 | 98.55 | 85.30 | 91.74 | 56.58 | 60.94 |
| **SWORD (Single)** | **97.20** | **98.82** | **86.95** | **92.72** | **58.25** | **62.55** |

*4.2.2 Single-Label Setting.* To validate the proposed method in the single-label setting, we compared it with recent single-label methods: SGGPoint [37] and SGFormer [22]. We adopted SGFormer as the backbone and replaced its original training objective with the proposed single-label SWORD-Margin loss in Eq. (32). We also evaluated the multi-label SWORD on the same label space under the single-label protocol by training it with the sigmoid-based multi-label SWORD objective and selecting the predicate with the highest sigmoid output for each directed object pair.

Table 3 shows that the multi-label variant already improves over SGFormer across all reported criteria, indicating that the dataset-internal object-relation geometry of SWORD is beneficial even without a single-label training objective. The single-label variant further achieves the best performance in every column, with a clearer advantage on predicate and relationship evaluation. Since the graph encoder is kept fixed, this gain is attributable to the supervision geometry introduced by SWORD-Margin rather than to an architectural change. This behavior is consistent with the structure of the objective. In the single-label setting, relation candidates are mutually exclusive, and the correct prediction must be selected jointly against other predicates and the no-relation state. SWORD-Margin in Eq. (32) matches this assumption by applying a cost-augmented softmax over $\mathcal{R}_\emptyset$, where incorrect labels are separated according to the scene-graph-grounded relation distance. Thus, semantically distant alternatives are discouraged more strongly, while nearby alternatives remain ordered by the learned graph geometry. In contrast, the multi-label variant predicts each predicate independently with sigmoid outputs. This formulation is natural when multiple relations may hold for the same object pair, but under one-label decoding the scores are not calibrated as mutually exclusive choices. As a result, generic or frequent predicates can remain competitive, and the null state is represented through induced probability mass rather than an explicit competing logit.

Figure 3 provides qualitative evidence for this interpretation. SGFormer often recovers part of the scene structure, but its errors tend to weaken the specificity of object-relation configurations. In the first scene, it replaces *blanket* with *towel* and maps bed-centered physical relations to the broader predicate *close by*. In the second scene, an object error from *shelf* to *fridge* leads to an implausible relation such as *clothes part of fridge*. SWORD preserves the main object identities and relation structure more consistently. When it differs from the reference, as in *pillow leaning against bed* or *cabinet hanging on wall*, the predicted relation remains compatible with the local geometry and is closer to the reference than a structurally inconsistent object-relation substitution.

Overall, these results consistently suggest that SWORD-Margin turns the scene-graph-grounded object and relation metrics into an effective training signal for single-label 3D scene graph prediction. It improves exact-label evaluation while preserving the intended behavior of SWORD: the reference graph remains the optimal solution, but graph-grounded near misses are distinguished from semantically and structurally inconsistent predictions.

## 4.3 Ablation Study

We analyzed the two key components of SWORD: the Wasserstein-based supervision objective in Sec. 3.3 and the scene-graph-grounded ground metrics in Sec. 3.2. All variants used the OCRL backbone and follow the same multi-label setting, dataset, tasks, and evaluation protocol as in Sec. 4.1. We only change either the training objective or the label distances used in the Wasserstein terms.

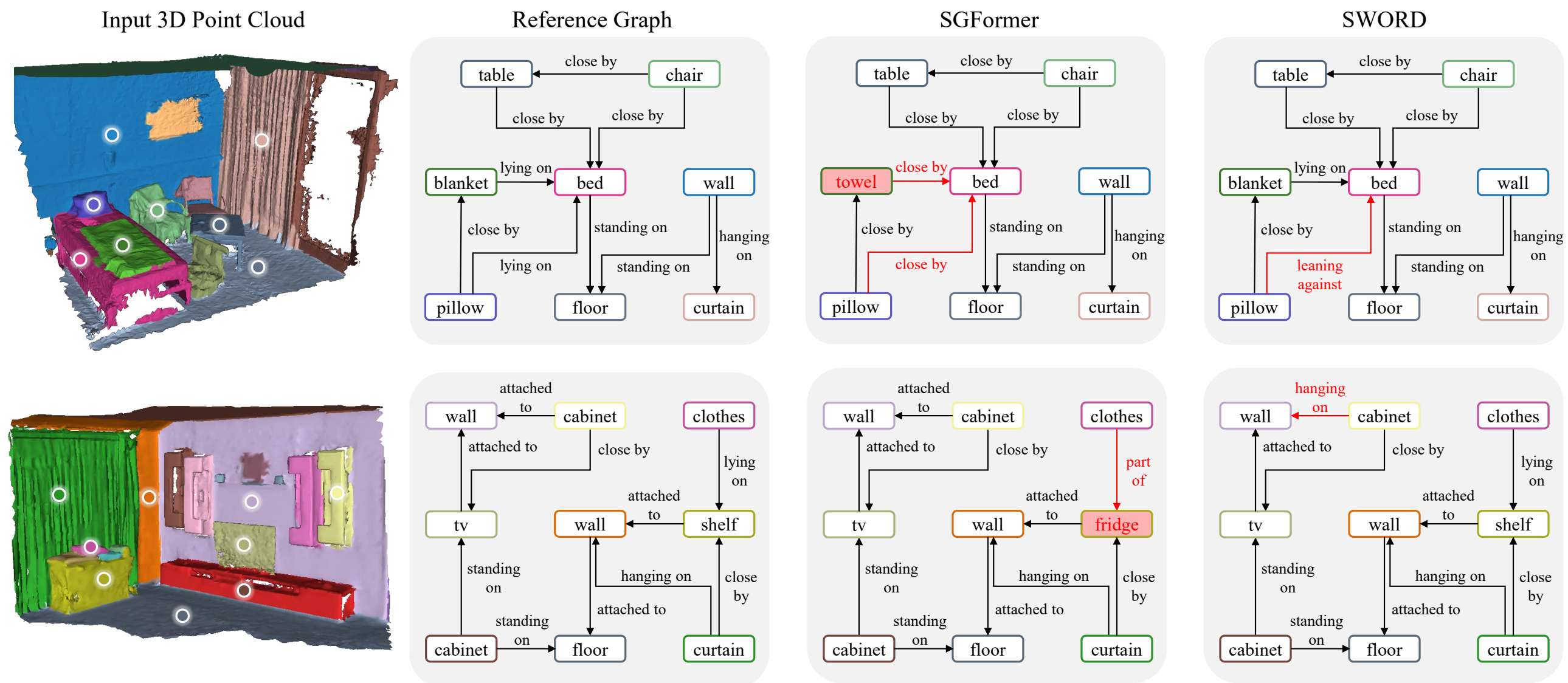

**Figure 3: Qualitative comparison in the single-label setting. Red nodes and edges denote deviations from the reference graph.**

For the ground-metric variants, we keep the proposed multi-label SWORD objective in Eq. (33) fixed and vary only the ground distance. *Uniform Dist.* uses a constant distance for all distinct labels, and *Text-Emb. Dist.* computes label distances from CLIP text embeddings [25] of object and relation names. *Base (w/o Context)* uses the base metrics $d_O^{\text{base}}$ and $d_R^{\text{base}}$ defined in Eqs. (14) and (16). For the objective variants, we keep the context-aware metrics $d_O$ and $d_R$ in Eqs. (18) and (20) fixed. *SWORD + CE/BCE* adds the original OCRL CE/BCE losses to Eq. (33) with unit coefficients and without tuning an additional mixing weight, whereas *SWORD (Ours)* uses the proposed SWORD objective without the CE/BCE terms.

Table 4 shows a consistent trend across object, predicate, triplet, PredCls, and SGCls evaluations. *Uniform Dist.* is the weakest among the ground-metric variants, indicating that merely replacing CE/BCE with a transport-style objective is not sufficient when the ground cost does not encode meaningful label geometry. *Text-Emb. Dist.* improves this behavior, suggesting that non-uniform semantic distances are useful. However, it remains below the graph-grounded metrics because language-level similarity between class names does not explicitly capture how object categories participate as relation endpoints, which predicates they co-occur with, or what structural roles they play in 3D scene graphs.

*Base (w/o Context)* further improves the results by constructing distances from dataset-internal object-relation statistics rather than external language priors. The full context-aware metric, *i.e.*, *SWORD (Ours)*, consistently performs best, showing that the additional context terms refine the label geometry beyond simple graph-token profiles. In particular, object labels are compared through their neighbor objects, incoming and outgoing relations, relation-neighbor contexts, and typed-degree profiles, while relation labels are compared through endpoint configurations, reverse relations, and local relation co-occurrence. This provides supervision that penalizes graph-semantically distant errors more strongly while allowing semantically plausible alternatives to remain closer in the induced metric space. The benefit is also reflected in mean-recall metrics, suggesting that the learned supervision is helpful beyond frequent dominant labels.

The objective comparison shows that SWORD + *CE/BCE* underperforms the proposed SWORD objective alone. CE/BCE imposes independent exact-label supervision on nodes and edges, whereas SWORD organizes prediction distributions according to the scene-graph-grounded distances. Adding CE/BCE can therefore dilute the distance-aware signal induced by the SWORD metric, especially when predictions should be shaped by semantic proximity rather than only by one-hot or multi-hot targets. Overall, the ablation confirms that SWORD requires both components: a metric-induced objective aligned with Eq. (33) and context-aware ground distances constructed from the internal structure of training scene graphs.

## 4.4 User Study

The quantitative comparisons in Sec. 4.2 show that SWORD improves standard exact-label metrics, even though it changes the supervision geometry rather than simply optimizing uniform CE/BCE-style local classification penalties. However, exact agreement with reference annotations has an inherent limitation: all incorrect labels are counted as errors, regardless of whether they are graph-grounded near misses or semantically distant substitutions. Therefore, the standard metrics can underrepresent improvements in plausibility, descriptiveness, and structural coherence. To evaluate these aspects, we conducted a blind A/B user study comparing OCRL [10] and SWORD.

We used two complementary settings. In the Graph-Graph setting, participants were given the reference graph and two predicted graphs, and selected the prediction that better satisfied *Semantic Closeness* and *Structure Preservation*. In the Scene-Graph setting, participants were given the reference 3D scene and two predicted graphs, and selected the prediction that better satisfied *Scene Descriptiveness* and *Relation Plausibility*. A total of 28 participants with normal or corrected-to-normal vision evaluated 20 validation samples. Each participant evaluated both settings, and each question was answered as a binary preference without ties; therefore, each criterion was computed over $28 \times 20 = 560$ pairwise judgments. The order of the two methods was randomized for each sample. All graphs were visualized with the same layout style, node and edge appearance, font, and label format to avoid method-unrelated visual bias. Table 5 reports the preference ratios for each criterion.

**Table 4: Ablation study of ground metrics and objectives in SWORD.**

| | Object | | Predicate | | Triplet | | PredCls | | SGCls | |
|---|---|---|---|---|---|---|---|---|---|---|
| *Ground Metric Variants* | R@5 | mR@5 | R@3 | mR@3 | R@50 | mR@50 | R@50 | mR@50 | R@50 | mR@50 |
| Uniform Dist. | 80.74 | 47.92 | 98.25 | 76.08 | 91.05 | 65.02 | 81.6 | 65.8 | 37.3 | 37.5 |
| Text-Emb. Dist. | 81.62 | 49.24 | 98.61 | 77.15 | 91.67 | 66.48 | 82.5 | 67.2 | 38.0 | 38.3 |
| Base (w/o Context) | 81.90 | 49.80 | 98.78 | 77.84 | 91.86 | 67.38 | 82.8 | 68.0 | 38.4 | 39.0 |
| *Training Objective Variants* | | | | | | | | | | |
| SWORD + CE/BCE Loss | 81.55 | 48.72 | 98.56 | 76.74 | 91.55 | 65.78 | 82.2 | 66.5 | 37.8 | 38.0 |
| **SWORD (Ours)** | **82.58** | **50.84** | **99.02** | **78.92** | **92.36** | **68.41** | **83.5** | **69.0** | **39.0** | **39.7** |

*Graph-Graph Comparison.* SWORD is clearly preferred over OCRL for both graph-based criteria. This result complements the exact-label evaluation in Sec. 4.2: while the standard metrics show consistent gains in strict object, predicate, and triplet prediction, the human comparison reveals a larger advantage in the semantic and structural quality of the predicted graphs. This gap is expected because exact-label metrics do not distinguish whether an error is a scene-consistent near miss or a semantically disruptive substitution. In contrast, SWORD explicitly orders object and relation labels by scene-graph-grounded distances, keeping predictions closer to the reference under exact-label uncertainty. The preference for *Structure Preservation* further suggests that SWORD better maintains coherent subject-predicate-object organization, rather than improving only isolated node or edge labels.

*Scene-Graph Comparison.* SWORD is consistently preferred when participants judge the predicted graphs directly against the reference 3D scene. The preference for *Scene Descriptiveness* indicates that SWORD better captures the object roles and overall scene organization perceived by humans. The preference for *Relation Plausibility* further suggests that its predicted relations are more compatible with the observed arrangement of objects. This is consistent with the design of SWORD: although its metric construction does not use external semantic priors or raw geometric features, it encodes endpoint contexts, local neighborhoods, typed-degree patterns, and relation co-occurrences from training scene graphs. These graph-grounded profiles guide the model away from semantically distant or structurally implausible alternatives.

**Table 5: User study results for Graph-Graph and Scene-Graph comparisons. Participants compared the baseline (*i.e.*, OCRL [10]) and ours under two questions for each setting.**

| Setting | Question | Baseline (%) | SWORD (%) |
|---|---|---|---|
| Graph-Graph | Semantic Closeness | 26.1 | 73.9 |
| | Structure Preservation | 24.1 | 75.9 |
| Scene-Graph | Scene Descriptiveness | 25.7 | 74.3 |
| | Relation Plausibility | 21.6 | 78.4 |

Overall, the user study highlights a key benefit that is only partially visible in conventional metrics. SWORD improves strict exact-label evaluation, but its larger advantage appears in the quality of the predicted graph as a semantic and structural description of the scene. These results support the central motivation of SWORD: replacing uniform penalties for all wrong labels with scene-graph-grounded distances mitigates annotation ambiguity while preserving exact ground-truth optimality.

## 5 Conclusion

We introduced SWORD, a scene-grounded Wasserstein object-relation distance for 3D scene graph prediction. By exploiting the fixed node and edge-slot correspondence in supervised protocols, SWORD defines an aligned graph distance, constructs object and relation ground metrics from training scene graph statistics, and provides distance-aware supervision beyond CE/BCE-style exact-label losses. This preserves the ground-truth graph as the unique zero-distance solution while distinguishing graph-grounded near misses from semantically distant errors. Experiments on 3DSSG show that SWORD improves object, predicate, relationship, and triplet prediction under both multi-label and single-label protocols. User studies further indicate that the predicted graphs are more semantically close, structurally preserved, scene-descriptive, and relation-plausible. These results show that SWORD improves not only exact-label benchmark accuracy but also perceived graph quality. Although SWORD demonstrates strong performance, like most 3D scene graph prediction scenarios, it cannot be applied off-the-shelf to settings where node correspondence, edge slots, or label vocabularies are not fixed. Future work will extend SWORD to partially aligned, open-vocabulary, and temporal scene graphs, while exploring geometric structure and perceptual 3D similarity for richer ground metrics [17, 18].